\documentclass[conference]{IEEEtran}
\IEEEoverridecommandlockouts
\usepackage{cite}
\usepackage{amsmath,amssymb,amsfonts}
\usepackage{algorithmic}
\usepackage{graphicx}
\usepackage{textcomp}
\usepackage{xcolor}
\usepackage{threeparttable}
\usepackage{multirow}
\usepackage{hyperref}
\hypersetup{%
 hidelinks,
 colorlinks=false}

\def\BibTeX{{\rm B\kern-.05em{\sc i\kern-.025em b}\kern-.08em
    T\kern-.1667em\lower.7ex\hbox{E}\kern-.125emX}}
\renewcommand{\baselinestretch}{0.99}
\begin{document}

\title{Cross-platform Learning-based Fault Tolerant Surfacing Controller for Underwater Robots \\
}

%\author{\IEEEauthorblockN{Yuya Hamamatsu, Maarja Kruusmaa, Asko Ristolainen}
%\IEEEauthorblockA{\textit{Department of Computer Systems} \\
%\textit{Tallinn University of Technology}\\
%Tallinn, Estonia \\
%email address or ORCID}}

\author{Yuya Hamamatsu$^{1}$, Walid Remmas$^{1}$, Jaan Rebane$^{1}$, Maarja Kruusmaa$^{1}$, Asko Ristolainen$^{1}$
\thanks{$^{1}$The authors are with the Department of Computer Systems, Tallinn University of Technology, Tallinn, Estonia
        {\tt\small (Yuya.Hamamatsu, Walid.Remmas, Jaan.Rebane, Maarja.Kruusmaa, Asko.Ristolainen)@taltech.ee}}
}

\maketitle

\begin{abstract}
In this paper, we propose a novel cross-platform fault-tolerant surfacing controller for underwater robots, based on reinforcement learning (RL). Unlike conventional approaches, which require explicit identification of malfunctioning actuators, our method allows the robot to surface using only the remaining operational actuators without needing to pinpoint the failures. The proposed controller learns a robust policy capable of handling diverse failure scenarios across different actuator configurations. Moreover, we introduce a transfer learning mechanism that shares a part of the control policy across various underwater robots with different actuators, thus improving learning efficiency and generalization across platforms. To validate our approach, we conduct simulations on three different types of underwater robots: a hovering-type AUV, a torpedo shaped AUV, and a turtle-shaped robot (U-CAT). Additionally, real-world experiments are performed, successfully transferring the learned policy from simulation to a physical U-CAT in a controlled environment. Our RL-based controller demonstrates superior performance in terms of stability and success rate compared to a baseline controller, achieving an 85.7 percent success rate in real-world tests compared to 57.1 percent with a baseline controller. This research provides a scalable and efficient solution for fault-tolerant control for diverse underwater platforms, with potential applications in real-world aquatic missions.
\end{abstract}

\begin{IEEEkeywords}
Fault-tolerant, Reinforcement learning, Cross-platform transfer learning, AUV, LSTM-PPO
\end{IEEEkeywords}

\section{Introduction}

Modern autonomous underwater vehicles (AUVs) have diversified in design and actuator geometry to optimize them for different purposes \cite{auvapp}, such as rudders, thrusters, and bio-inspired mechanisms. They require specialized control strategies for each design \cite{soft1} \cite{soft2} \cite{bios}. Developing a robust controller capable of handling these diverse actuator configurations becomes particularly critical when accounting for potential actuator malfunctions. One of the fundamental tasks in almost all underwater missions is surfacing. Even if some actuators have problems, the robot control must be robust enough to be able to return using the available actuators. This study addresses the surfacing task in these malfunctioning situations. It also allows for sharing part of the control policy model using transfer learning \cite{tr1} \cite{tr2} across multiple different actuator designs.

\begin{figure}[tbp]
\centerline{\includegraphics[width=0.9\linewidth]{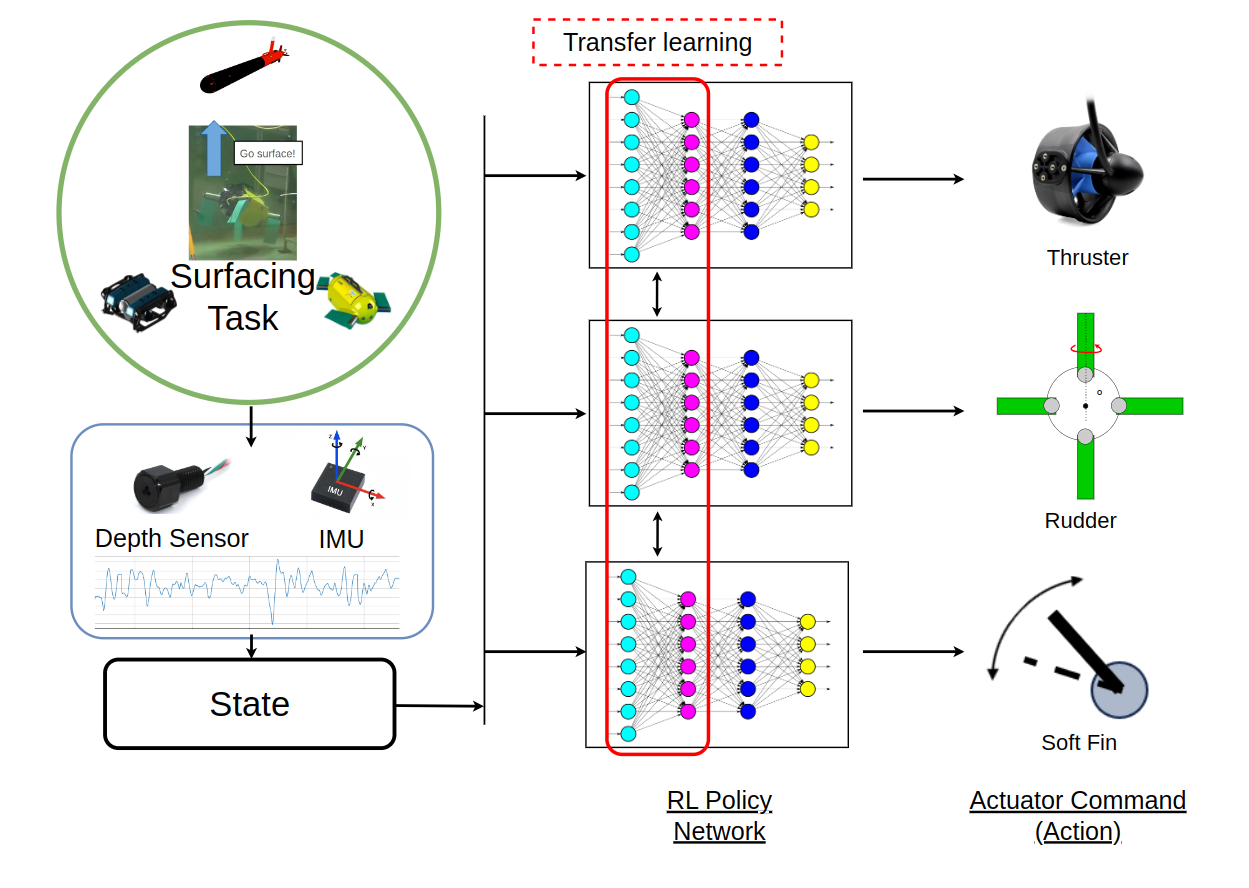}}
\caption{Overview of the proposed method. The task to be accomplished is that when at least one or more actuators are in a failed state, the robot will use the sensor information to surface using the actuators that are operational. In this case, the identification of the failed actuator is not required. In addition, by transferring the first layer of the policy network, part of the RL model is shared between multiple different platforms.}
\label{task}
\end{figure}

Fault-tolerant control strategies are basically designed as a specific controller for each malfunctioning state \cite{nlFT1} \cite{nlFT2}. This approach requires the identification of malfunctioning actuators \cite{FTquad} \cite{FTquad2} \cite{FTUAV1}.
To overcome these limitations, we propose a novel control strategy based on reinforcement learning (RL) that enables underwater robots to perform the surfacing task without explicitly identifying the faulty actuators. Our RL-based approach learns a robust policy capable of handling a variety of failure scenarios and actuator configurations. Unlike conventional methods \cite{walid}, our approach does not require the identification of the malfunctioning actuator. The proposed learning-based method also covers a combination of multiple actuator malfunctions with a single controller. Furthermore, it facilitates the transfer of parts of the learned RL policy between different robot platforms with various actuator types for the same task. This policy transfer capability improves learning efficiency and supports the development of more generalizeable control strategies. This is achieved by sharing the weights of the actor and critic network inside the RL agent between each platform. This transfer of learning policies increases the flexibility and robustness of our approach. Fig. \ref{task} shows the overview of the proposed method.
\begin{figure*}[t]
\includegraphics[clip, width=18cm]{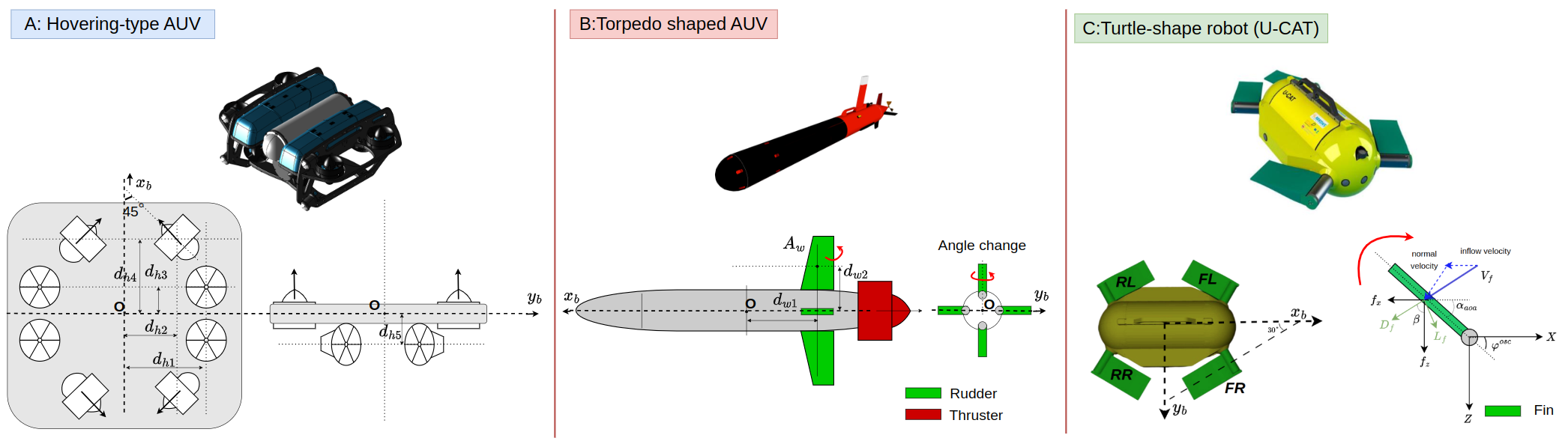}
\caption{Three different platform models and actuator configurations. (A) Hovering type AUV with 8 thrusters. 8 thrusters are arranged symmetrically and the arrow on the thruster indicates the positive direction of output. (B) Torpedo-shaped AUV. Green area indicates the rudders, which can change angle as direction of red arrows. Red indicates thrusters for thrust in the surge direction, and arrows indicate positive direction of output. (C) Turtle-shaped robot (U-CAT). The green part shows the fins, which are paddled to generate thrust. Dynamics of the fin is shown in the lower right.}
\label{ucat}
\end{figure*}
We validate our proposed method through simulations on three different types of underwater vehicles: a hovering-type AUV \cite{hover} with eight thrusters, a torpedo-shaped AUV \cite{torpedo1} \cite{torpedo2} equipped with four rudders and a thruster, and a turtle-shaped fin-actuated robot (U-CAT) \cite{ucat0}.  

Furthermore, to illustrate the practical applicability, we conducted Sim2Real transfer experiments with U-CAT in a swimming pool, successfully deploying the controller trained entirely in simulation in real-world scenarios \cite{s2rreview}, \cite{s2r1}, \cite{s2r2}, \cite{s2r3}. This approach eliminates the risk of hardware damage due to intensive testing of failure modes in actual environments and allows mechanical repetition of random failure scenarios.

% The main contributions of this research are the following:
% \begin{quote}
%  \begin{itemize}
% \item \textbf{Unified Control Strategy for Multiple Actuator Failures}:  We introduce a reinforcement learning-based control strategy capable of managing multiple actuator failures across different underwater robot configurations without requiring explicit fault identification.

% \item \textbf{Enhanced Generalizability through Policy Transfer}:  We demonstrate that portions of the RL policy can be transferred across robots with different actuator configurations, provided they perform similar tasks. This enables the development of more generalizable controllers that can be adapted to a variety of underwater robotic systems.

% \item \textbf{Simulation-Based Fault Reproduction Using Sim2Real Transfer}: By training in simulated environments, we can replicate various actuator failure scenarios, enabling a comprehensive evaluation of the controller's robustness in handling diverse malfunctions. This approach allows for a safe and efficient exploration of failure modes without risking real hardware.
% \end{quote}

\section{Robot Dynamics And Kinematics}
This section describes the platforms for the validation of our method, its simulation model, and the actuation rules for 3 different platforms. 

For all robots, the dynamics of the body frame used in the simulation is represented by the Fossen model \cite{fossen}:

\begin{equation}
M \dot{\nu}+ C(\nu) v+D(\nu) v+g(\eta)=\tau
\end{equation}

where the vector $g$ represents the gravitational/buoyancy forces and moments. The term $\tau$ represents the forces generated by the actuators. $M$ represents the vehicle inertia matrix, $C$ represents the Coriolis matrix, and $D$ denotes the damping matrix. The robot motion is simulated by reflecting the forces and torques generated by the actuators. The vector of linear and angular velocities in the robot frame is described as $\nu=[u, v, w, p, q, r]^T$, and the pose of the robot in the earth-fixed frame is $\eta = [x, y, z, \theta, \phi, \psi]^T$. 
The following subsections describe each specific input/output, dynamics, and failure scenarios of the AUV actuators for each design. 

\subsection{Hovering type AUV with 8 thrusters}

A hovering type AUV \cite{hover} is actuated with 8 thrusters, four in the heave direction, and the other four installed at an angle of 45 degrees to the sway and surge direction, as shown in Fig.2.A. The hovering AUV dynamic simulation model was developed following the Heavy Configuration of the Blue ROV \cite{blue} and the numerical parameters of the model are set according to \cite{bluedynamics}. Each thruster is independently controlled and takes force as input. They generate the force in the direction the thruster is installed. For the hovering AUV, failures are randomly assigned in simulation for up to 7 thursters, with at least one thurster in the heave direction being faultless. The thrusters in the fault condition will always output a 0 N force regardless of the command input.

\begin{figure*}[t]
\includegraphics[clip, width=18cm]{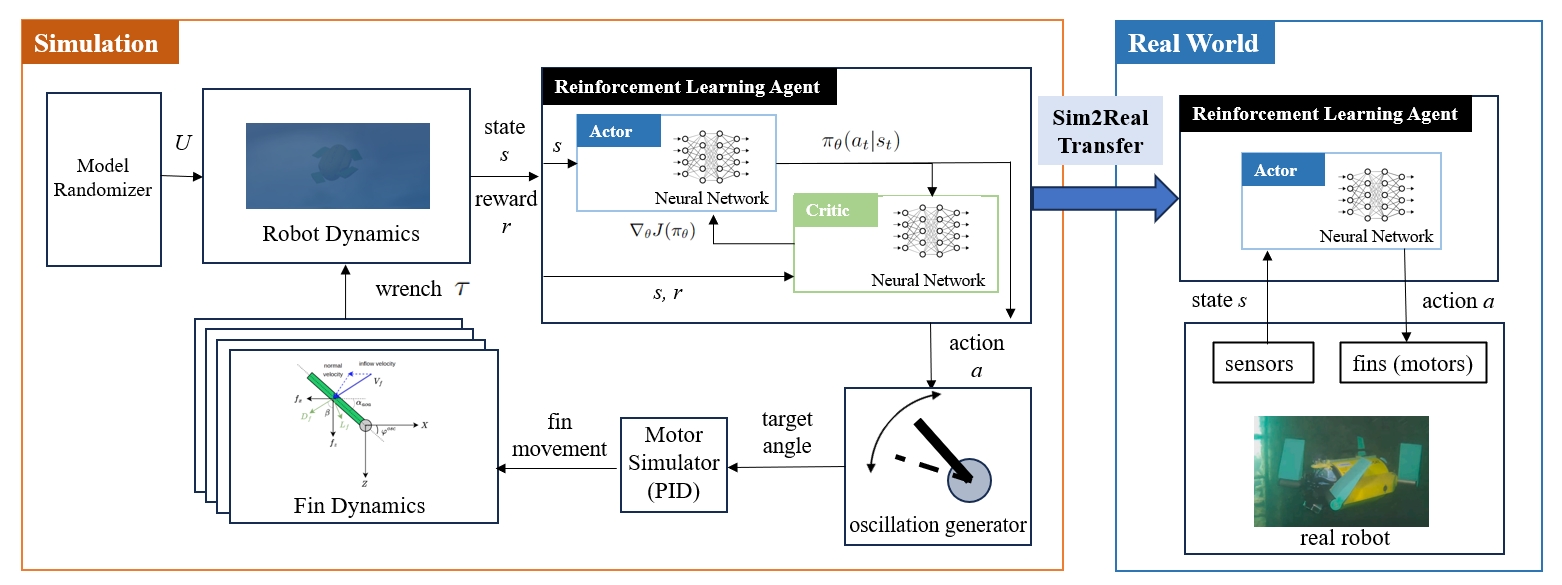}
\caption{Schematic diagram of an Actor-Critic based RL controller for turtle-shaped robot (U-CAT) using simulated dynamics. The left side shows the procedure for training the controller in the simulator, and the right side shows how to apply the controller to the real robot (Sim2Real).}
\label{rl}
\end{figure*}

\subsection{Torpedo-shaped AUV}

The second simulated robot is a torpedo-shaped AUV with four independently actuated rudders and a single thruster that generates thrust in the surge direction. The actuator dynamics is based on the models described by \cite{torpedo1} \cite{torpedo2}. Each rudder is positioned at 90-degree intervals around the vehicle and their angles can be independently adjusted allowing the generation of both lift and drag forces for control of the vehicle's orientation. The rudders installed vertically to the body rotate in the yaw direction, and horizontal ones rotate in the pitch direction, as shown in Fig. \ref{ucat}.B.

The forces generated by each rudder include both lift, which is perpendicular to the direction of the fluid flow, and drag, which acts parallel to the flow:

% The lift force \(L_i\) produced by each wing is dependent on the relative velocity of the fluid and the angle of attack, and the drag force \(D_i\) opposes the motion through the fluid and is also a function of the relative velocity of the fluid and the angle of attack given by the following.
\begin{equation}
\begin{aligned}
L_i = \frac{1}{2} \rho V^2 A C_{L_i}(\alpha_i) \\
% \end{equation}
% \begin{equation}
D_i = \frac{1}{2} \rho V^2 A C_{D_i}(\alpha_i)
\end{aligned}
\end{equation}

where $L_i$ and $D_i$ are the lift and drag force, $\rho$ is the fluid density, $V$ is the relative velocity of the fluid, \( A \) is the area of the rudder and \( C_{L_i}(\alpha_i) \), \( C_{D_i}(\alpha_i) \) are the lift and drag coefficient as a function of the angle of attack \( \alpha_i \). These coefficients are numerically calculated as follows \cite{rudder}.

\begin{equation}
\begin{aligned}
C_{L_i} = 0.13058\alpha + 0.051143\alpha |\alpha| \\
C_{D_i} = 0.0015587\alpha^2 + 0.058202
\end{aligned}
\end{equation}

% While the thruster generates linear propulsion to drive the AUV forward, the lift and drag generated by the rudders are crucial for attitude control. 
% Adjusting wing angles facilitates the generation of the necessary forces and moments, allowing the AUV to efficiently control pitch and roll, maintaining the desired attitude. 
The control input is a five-dimensional array of target angles (radians) for each rudder and the target output of the thruster (N).

In our study, one to three rudders are randomly set to malfunction, but at least one pitching rudder is kept as operational. The faulty rudder maintains the initialization angle (0 degree).

\subsection{Turtle-shaped robot (U-CAT)}

U-CAT is an Autonomous Underwater Vehicle (AUV) inspired by turtle locomotion \cite{ucat2}. It moves using four soft fins. Unlike conventional thruster actuation for underwater robots, fin propulsion has substantially more nonlinear coupled dynamics. Fig. \ref{ucat}.C shows the fin configuration. 

Each fin can be controlled independently to generate thrust by modulating the amplitude $A^{amp}$, the zero direction of oscillation $\phi^{c}$, the oscillation rate $\omega^{o s c}$ and the phase offset $\varphi_{o f f}^{o s c}$. The oscillatory movement for each fin in time $t$ is described by:
\begin{equation}
\label{osc}
\varphi^{o s c}(t)=A^{amp} \sin \left(\omega^{o s c} t+\varphi_{o f f}^{o s c}\right)+\phi^{c}
\end{equation}

\noindent where $\varphi^{o s c}(t)$ is set as the next target angle of the fin.  
% A PID controller controls the motion for each motor driving the fin, both in simulation and in the real robot.

To calculate the generated body forces $\tau$ in simulation, the dynamic model based on the rigid fin model was used \cite{fin}.
Each fin generates horizontal $f_x$ and vertical $f_z$ forces (N) relative to the angle of the fin, as shown in Fig. \ref{ucat}.C. These forces are expressed as follows.

\begin{equation}
\begin{aligned}
& f_x=D_f \sin (\beta)+L_f \cos (\beta) \\
& f_z=-L_f \sin (\beta)+D_f \cos (\beta)
\end{aligned}
\end{equation}

Here, $\beta$ denotes the direction of the flow with respect to the fin, where $D_f$ and $L_f$ are the drag and the lift forces defined in \cite{fin}.
In this study, up to 3 fins are randomly set to malfunction. The failing actuators maintain the same initial angle regardless of the command.

A baseline controller was also used to control the robot to compare its performance in real-world experiments. Several advanced controllers for U-CAT have been developed and tested in previous works \cite{ucat_depth}. However, none of these controllers considers unidentified malfunctioning fins. Therefore, we used the PID controller that generated force vertically to the robot's body frame, and the feedback is calculated as the depth error. The output of the controller is converted to an oscillation profile using the model described in \cite{ucat_inverse}.

\section{Reinforcement learning based fault-tolerant surfacing controller}
Fig. \ref{rl} shows an overview of our RL based fault-tolerant surfacing control architecture. As shown in Fig. \ref{rl}, the control policy is first trained in the simulator and then transferred to the real robot. In this study, this Sim2Real transfer is only performed using the U-CAT case.

\subsection{Learning algorithm}
RL involves an agent interacting with an environment based on the current state $s_t$ , selecting an action $a_t$, executing it, and observing the next state \( s_{t+1} \) along with the reward $r_{t+1}$. 

Several learning algorithms have been proposed for optimal policy convergence. The Proximal Policy Optimization (PPO) algorithm represents a significant advance aimed at streamlining the optimization of policy functions \cite{ppo}. PPO differentiates itself through the application of Policy Gradient methods \cite{pg} to update the actor network. 

PPO offers a significant advantage in managing the complexity and unpredictability of robot control, especially in situations where the failure of specific actuators cannot be directly pinpointed. One of the key features is its policy update mechanism, which uses a clipping function to limit the size of policy updates. This approach effectively prevents the policy from converging too much on actions that appear optimal based on limited specific fault assignments. The ability of the PPO to maintain a balance between exploration and exploitation makes it particularly well suited for handling scenarios involving random actuator failures. By limiting the influence of any single observation, PPO ensures that the control policy remains versatile and adaptive, capable of handling a wide range of operational anomalies without over fitting to specific fault assignment.
We improve the PPO framework by incorporating long- and short-term memory (LSTM) networks \cite{rnn}, which are adept at capturing crucial temporal dependencies for dynamic control scenarios. 
%This modification is particularly beneficial for tasks that require precise path-following capabilities, such as in unmanned surface vehicles \cite{asv}, where our adapted framework has shown superior performance in managing complex navigation tasks.

\subsection{Training procedure}
 All controller training procedures are carried out in simulation based on the physics models following Section II. The controller is trained to set the optimal observations, actions, and rewards for each of the established states as the following subsections.

\subsubsection{Episode initiation}
In the beginning of each episode, several actuators are randomly set to a fault state according to the conditions described in Section II. The RL agent is able to send actions to faulty actuators, but they do not respond or move. The robot does not receive information about the faulty actuators or any feedback from them. The differences from the real world and the simplified model are complemented by a training process using domain randomization techniques that add random values to the simulation parameters \cite{domain}. The dynamics parameters are added as random noise using a uniform distribution with a range from $a_i$ to $b_i$ as $U_i (a_i,b_i)$ at the beginning of each episode. Each episode ends and is reset when the robot successfully surfaces or reaches the time limit. Then these initiation processes are performed again, and the new episode starts.

\subsubsection{State}
%Action policy is chosen depending on the state values not only in training but also in real experiment.
%Therefore, these values had to be available not only in the simulation but also for the actual robot using sensor information. 
The state includes the vertical velocity in the world coordinate system $v_{z}^w$, derived from the pressure sensor, the pose represented by a quaternion for the attitude, the angular velocity using Euler angles, all observed by the Inertial Measurement Unit (IMU), with an Extended Kalman Filter (EKF) \cite{ekf} applied to reduce noise.
% The velocity in the depth direction $v_{z}^w$ in the world coordinate system calculated from the derivation of the pressure sensor was used as one of the state values. In addition, state values are also indicated by pose, including attitude described by quaternion and angular velocity described by Euler orientation. These values are observed by the Inertial Measurement Unit (IMU). When the attitude is calculated, an Extended Kalman Filter (EKF) \cite{ekf} is applied to reduce the measurement noise.

\subsubsection{Action}
Actions are determined based on the inputs of the robot actuators, thus each platform has a different dimension of the action space. The actions are the outputs of the actuators and are described in each subsection of Section II for each platform.
\subsubsection{Reward}
% \begin{equation}
% r_{step}^{vel} = \frac{v_{z}^w}{n_{active}}
% \end{equation}
$v_{z}^w$ and the uprightness reward is calculated to evaluate the robot's stability. The uprightness reward is calculated as
$\boldsymbol{\hat{z}}^T[0,0,1]$. $\hat{z}$ represents the direction vector of the robot's z axis in world coordinates. The uprightness reward helps the robot to balance between maintaining stability and taking actions needed to surface, at the same time preventing aggressive movements that could change its attitude too much. This reward gets closer to 1 when the robot is upright and gets closer to 0 when it leans far. This helps to keep the robot stable while still allowing it to perform the necessary movements for surfacing. Then, the total immediate rewards obtained at each time step can be expressed as follows:

\begin{equation}
r_{step}= k_1v_{z}^w + k_2\boldsymbol{\hat{z}}^T[0,0,1]
\end{equation}
where $k_1$ and $k_2$ are weight parameters.

The goal rewards $r_{goal}$ are calculated to evaluate the rewards for each episode as a whole. They are added when the robot has successfully surfaced or reached the time limit.

\begin{equation} r_{goal}= k_3\sqrt{(x_{f} - x_{s})^2 + (y_{f} - y_{s})^2} + k_4 \frac{(T_{l} - t)}{T_{l}} + k_5 \end{equation}

where $x_{f}, y_{f}$ denotes the final position of surfacing of the robot in the world coordinate system, and $x_{s}, y_{s}$ represents the initial position of starting of the robot, set to $(x_{s}, y_{s}, z_{s})$. The variables $k_3$, $k_4$, and $k_5$ are weight parameters. The execution time limit for each trial is $T_{l}$, and the time taken to complete the task is $t$. If the time in an episode reaches $T_{l}$, the episode ends and the RL agent does not receive the goal reward $r_{goal}$.

\subsection{Cross-platform transfer learning}
To improve the generalization and transferability of the policy network across different underwater robot designs, we partially standardize the internal structure of the LSTM networks used in both the actor and critic models. This makes some parts general and reduces the amount of learning required to accomplish a task. Specifically, we share a portion of the LSTM layer weights between the actor and critic networks across different robot designs. In this study, the observation space has the same number of dimensions for all platforms because it is assumed to be observed by the same sensor. Therefore, the neural network has the same structure except for the final layer that sends commands to actuators. It is implemented by directly copying the weights of other platforms that have already been trained. These shared weights are allowed to be updated during training. The shared LSTM layers process essential sensor data, such as depth and attitude measurements, which are critical environmental observations in underwater robotics. These sensors are inexpensive and commonly used across various robotic platforms, facilitating seamless integration. By sharing parts of the LSTM layers, we enable efficient knowledge transfer across different robot architectures. In contrast, the layers closest to the output of the policy network require optimization specific to the individual characteristics of actuators and control mechanisms of each robot. Given the diversity in actuator designs and control strategies, weight sharing is avoided in these adjacent output layers. Instead, these layers are trained independently to ensure that the policy network can adapt to the unique dynamics and physical constraints of each robot. This approach ensures a balance between generalization, where shared LSTM layers promote knowledge transfer, and robot-specific optimization, where the output-proximal layers specialize in precise control for each robot.

\section{Results and discussion}
\subsection{Training on simulator}

% \begin{table}[t]
%     \caption{RL parameters}
%     \label{tab:rlparam}
%     \begin{center}
%     \begin{tabular}{|c|c|}
%     \hline
%     \textbf{Parameter} & \textbf{Value} \\
%     % \hline
%     % \multicolumn{2}{|c|}{General settings parameters}  \\
%     % \hline
%     % Total Timestep & 300000 \\
%     % Limit time ($T_{limit}$) & 400 sec \\
%     % Step frequency & 2 Hz \\
%     % \hline
%     % \hline
%     % \multicolumn{2}{|c|}{Robot and environment parameters}  \\
%     % \hline
%     % Initial depth & 5.0 m \\
%     % Goal depth & 0.2 m \\
%     \hline
%     \hline
%     \multicolumn{2}{|c|}{Reward weights}  \\
%     \hline
%     Velocity weight ($k_1$) & 4.0 \\
%     Upright weight ($k_2$) & 0.4 \\
%     Goal distance weight ($k_3$) & -4.0 \\
%     Goal time weight ($k_4$) & -20.0 \\
%     Goal reward ($k_5$) & 500.0 \\
%     \hline
%     \hline
%     \multicolumn{2}{|c|}{PPO and LSTM parameters}  \\
%     \hline
%     Learning rate & 0.0003 \\
%     Discount factor ($\gamma$) & 0.99 \\
%     Clipping parameter ($\epsilon$) & 0.2 \\
%     Number of LSTM layers & 3 \\
%     Number of LSTM units & 64 \\
%     \hline
%     \end{tabular}
%     \end{center}
%     \end{table}

\newcommand{\spheading}[2][5em]{{\parbox{#1}{\raggedright #2}}}
\begin{table}[t]
    \caption{Training parameters}
    \label{tab:rlparam}
    \begin{center}
    \begin{tabular}{|c||c|c|}
    \hline
    \textbf{} & \textbf{Parameter} & \textbf{Value} \\
    % \hline
    \hline
    \multirow{5}{*}{\spheading{Reward weights}} 
        & Velocity weight ($k_1$) & 4.0 \\
        & Upright weight ($k_2$) & 0.4 \\
        & Goal distance weight ($k_3$) & -4.0 \\
        & Goal time weight ($k_4$) & -20.0 \\
        & Goal reward ($k_5$) & 500.0 \\
    \hline
    \multirow{5}{*}{\spheading{PPO and LSTM parameters}} 
        & Learning rate & 0.0003 \\
        & Discount factor ($\gamma$) & 0.99 \\
        & Clipping parameter ($\epsilon$) & 0.2 \\
        & Number of LSTM layers & 3 \\
        & Number of LSTM units & 64 \\
    \hline
    \end{tabular}
    \end{center}
\end{table}

\begin{figure}[tbp]
\centerline{\includegraphics[width=0.95\linewidth]{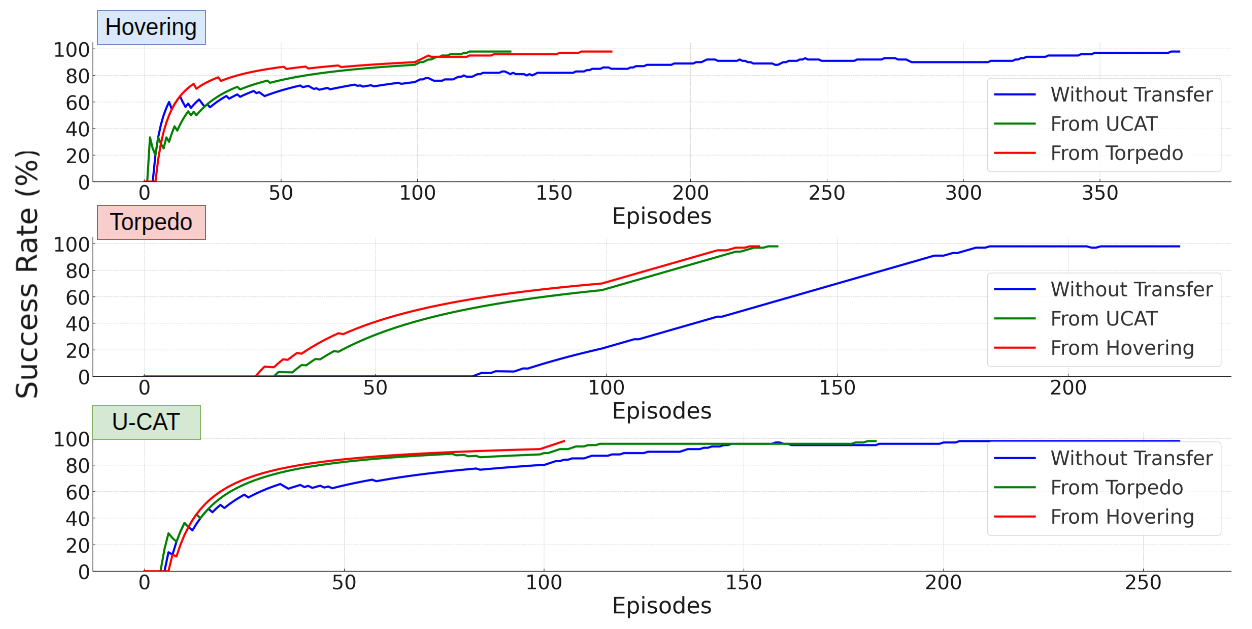}}
\caption{Success rate transitions in the last 100 trials during training of the model (Vanilla and first layer transfer).}
\label{tr}
\end{figure}
% From top to bottom, results for Hovering type, Torpedo-shaped, U-CAT. Blue line indicates the case without transfer learning, and the other colors indicate the results of 1st layer transferring from a trained model of another robot.

\begin{table}[t]
\begin{minipage}{\linewidth}
\caption{Transfer Learning Efficiency}
\label{tab:tr}
\begin{center}
\begin{tabular}{|c|c||c|c|c|}
\hline
\multicolumn{2}{|c|}{Transfer Model \textbackslash Robot} & Hovering & Torpedo & U-CAT  \\
\hline
\hline
\multirow{2}{*}{Hovering} & Layer 1 &   \multirow{2}{*}{1.0 (382)\footnotemark[1]} &  \textbf{0.60(136)} & \textbf{0.41(108)}\\
                          &Layer 1,2 &        &       0.85(194) & 1.13(298)\\
\hline
\multirow{2}{*}{Torpedo} & Layer 1 &      \textbf{0.45(174)} & \multirow{2}{*}{1.0(227)\footnotemark[1]} & \textbf{0.61(140)} \\
                         & Layer 1,2 &    1.87(717)  &     & 0.78(205)\\
\hline

\multirow{2}{*}{U-CAT} & Layer 1 &    \textbf{0.36(136)}&       \textbf{0.71(186)}&    \multirow{2}{*}{1.0(262)\footnotemark[1]} \\
                       &Layer 1,2 &   1.31(502) &       0.83(189)&         \\
 \hline
\end{tabular}
\end{center}
\footnotemark{They indicate reults without transfer learning (vanilla model)}
\end{minipage}
\end{table}

The RL parameters are shown in Table \ref{tab:rlparam}. The malfunctioning actuators were randomly chosen in the beginning of each episode and the robot was trained to reach the surface from an initial depth of 5.0 m. Each episode ended when the robot reached the surface or when the time limit (Hovering: 40s, Torpedo: 100s, U-CAT: 75s) was reached.  The training was finished once the robot achieved a 99 percent success rate in surfacing within the time limit, based on the last 100 episodes. For each robot, a baseline "vanilla" model was trained without transfer learning. Transfer learning was then applied by transferring each layer of LSTM parameters of both the actor and the critic networks from the other platform's vanilla models to accelerate the learning process. Two types of transfer learning were performed for each combination of platforms: transferring weights for the first layer only and transferring both the first and second layers. The progression of the success rate of the "vanilla" model and the transferred model in the first layer during training for each robot is shown in Fig.\ref{tr}, and the efficiency achieved by transfer learning is summarized in Table \ref{tab:tr}. Table \ref{tab:tr} shows the ratio of episodes required to achieve a 99 percent success rate in the last 100 episodes, relative to the number of episodes needed in the vanilla model (which is set to 1.00 for comparison). The values in parentheses indicate the actual number of episodes required to reach this success rate.

The first LSTM layer transfer learning significantly accelerated the training process for each model. However, both first- and second-layer transitions do not always contribute to the efficiency of robot policy learning. Some combinations took more episodes than without transfer learning. The reason might be that the  closer the layer is to the output, the more the weights reflect  characteristics of the specific actuator.

% Fig. \ref{sim} shows the mean reward values for each algorithm. The PPO-based method received a higher mean reward than the SAC-based one. In addition, LSTM-PPO quickly converged compared to the basic PPO. Therefore, the trained LSTM-PPO based controller was transferred to the real robot in the following sections.

\subsection{Real-world experiments}

% \begin{table}[t]
% \begin{minipage}{\linewidth}
%     \caption{Robot hardware specification}
%     \label{tab:robotparam}
%     \begin{center}
%     \begin{tabular}{|c|c|}
%     \hline
%     \textbf{Parameter} & \textbf{Value} \\
%     \hline
%     Weight & 18.1 kg \\
%     Motors & Maxon Motor 272763 \\
%     Motor controller & Maxon Motor EPOS2 \\
%     IMU & Microstrain 3DM-CX5-IMU  \\
%     Pressure sensor & Gems 3101P-0016G-01-B-000 \\
%     Camera resolution\footnotemark[1] & 1280 x 720 (pixels) \\
%     Camera field of view\footnotemark[1] & 69.4 x 42.5 (degree) \\
%     Action frequency & 2 Hz \\
%     Attitude estimation frequency & 10 Hz \\
%     Depth estimation frequency & 100 Hz \\
%     \hline
%     \end{tabular}
%     \end{center}
% \footnotemark[1]{This sensor was used only for the performance evaluation. Not used for the proposed method itself.}
% \end{minipage}
% \end{table}

\begin{figure}[tbp]
\centerline{\includegraphics[width=0.9\linewidth]{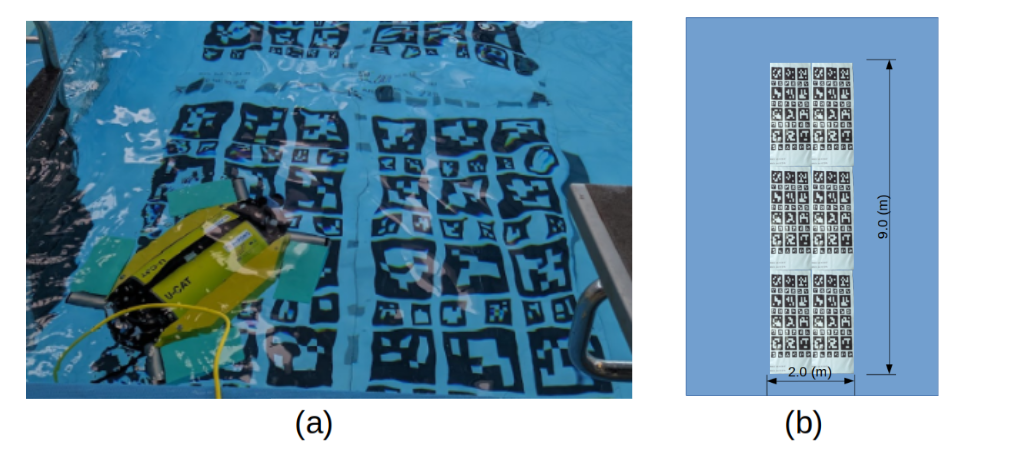}}
\caption{(a) The test setup in a swimming pool (Keila, Estonia). (b) The ArUco markers and dimensions of the workspace.}
\label{experiment}
\end{figure}

The controller trained in the simulations was also transferred to the real U-CAT robot. The vanilla model was used for the real-world experiments. The robot was deployed in a pool with a depth of 1.65 m. It was equipped with a pressure sensor to measure depth and an IMU sensor to measure attitude.
As shown in Fig. \ref{experiment}, the robot was equipped with a camera to capture the robot's trajectory using ArUco markers \cite{ar} placed at the bottom of the pool. The valid operating area is defined as the area where at least one marker is within the camera's field of view.

The 14 possible combinations of malfunctioning actuators were defined as one set. Trials using the RL controller are carried out in two sets (28 trials total), and trials with the baseline controller were carried out in one set (14 trials). The baseline controller was explained in Section II.C.
Trials were marked as successful if the robot reached the surface in 60 seconds. When the robot left the work area for more than 5 seconds, these trials were marked as failure. The buoyancy of the robot was adjusted to be slightly negative.

\begin{figure*}[t]
\includegraphics[clip, width=18cm]{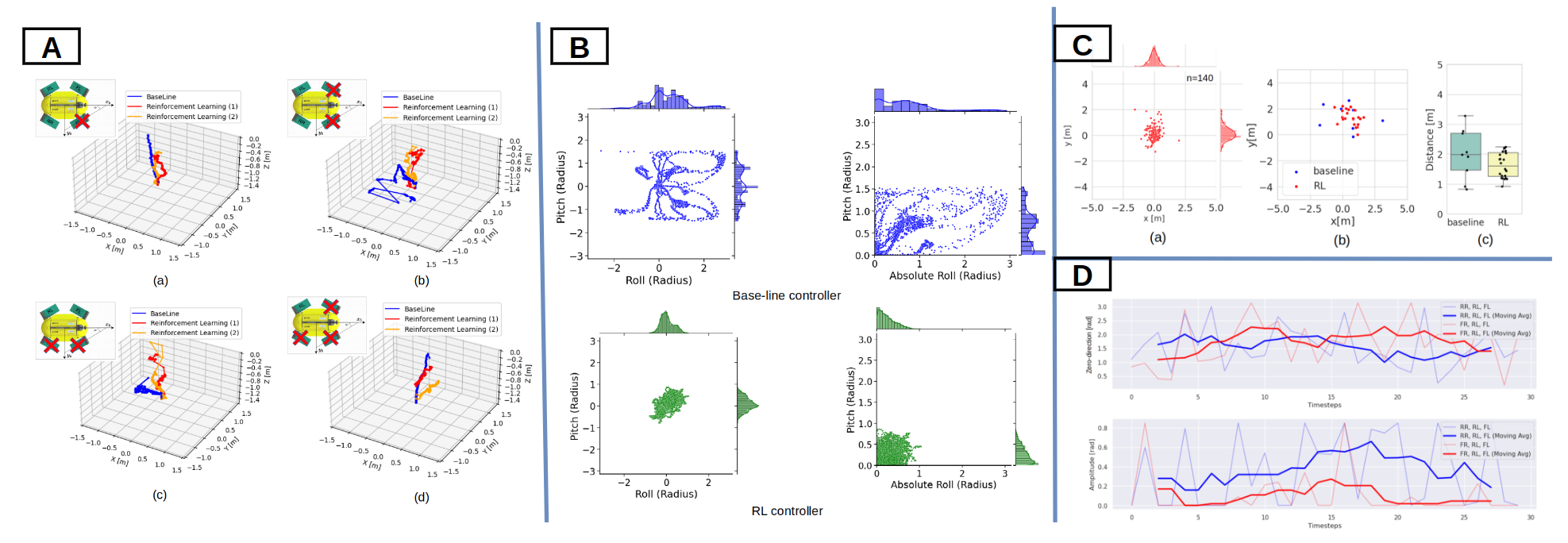}
\caption{A: Trajectories captured by camera using ArUco markers. Upper left insert shows malfunctioning actuator combinations of each trial. (a) FR, (b) FR/FL (c) FR/RR, (d) FR/RR/FL. In the scenario (b), the baseline controller was not success to reach water surface. B: Distributions of roll and pitch attitude estimated by the IMU sensor. The top row shows when the robot used the baseline controller, and the bottom row shows when the robot used the proposed RL controller. The left column are raw values of the attitudes, and the right column are absolute values. C: (a) Surfacing positions of successful trials using RL controller in simulation after training. (b) Positions of successful surfacing in real experiments . Blue points shows the case of using baseline controller, and red shows the case of using the RL controller. (c)  the distribution of distances from x,y position of the starting position in each controller in the real experiment. D:The figures show the action values of the functioning actuator. Red indicates failed cases (FR, RL, FL) and blue indicates successful cases (RR, RL, FL). (FR, RL, FL) and (RR, RL, FL) indicate the combination of faulty actuators. Thin lines indicate the raw value and thick lines indicate the moving average. The window size for the moving average is set to 5.}
\label{ex}
\end{figure*}

\begin{table}[t]
\begin{minipage}{\linewidth}
    \caption{Experiment results}
    \label{tab:experimentresult}
    \begin{center}
    \begin{tabular}{|c|c|c|}
    \hline
    \textbf{Description} & \textbf{RL controller} & \textbf{Baseline controller} \\
    \hline
    Success rate & 24/28 (85.7\%) & 8/14 (57.1\%) \\
    \hline
     & & (RR), (RL), \\
    Failed combinations & (FR, RL, FL)\footnotemark[1] & (FR, FL), (RR, RL), \\
     & (FR, RR, RL)\footnotemark[1] &  (LR, FL), (FR, RR, RL) \\

    \hline
    \end{tabular}
    \end{center}
\footnotemark[1]{Failed combination of faulty actuator are same in both sets.}
\end{minipage}
\end{table}

The experimental results are summarized in Table \ref{tab:experimentresult}. The robot successfully reached the surface in 24 out of 28 trials when the proposed RL controller was applied and in 8 out of 14 trials when the baseline controller was applied. 
The unsuccessful trials had the same combinations of malfunctioning actuators in both sets. 
Fig. \ref{ex}.A shows the estimated trajectories captured by the camera using ArUco markers in the represented trials. 
Fig. \ref{ex}.B shows the roll and pitch attitude distributions estimated by the IMU sensor.  
The robot that uses the baseline controller showed a large variation in trajectories, the standard deviation of the roll angle was 0.683 (rad), and the pitch angle was 0.419 (rad). On the other hand, the RL controller performed more stably, the standard deviation of roll angle was 0.192 (rad) and the pitch angle was 0.183 (rad). 
According to these results, the robot using the baseline controller showed a large variation in attitude, while the robot using the proposed RL controller exhibited a more stable behavior.
In all the failed trials when the robot used the baseline controller, the robot thrusted too actively and went out of the valid operating area. On the other hand, all the failed trials when the robot used the RL controller, the robot thrusted too passively to the surface using one fin, and therefore could not reach the surface within the time limit. We discuss the details of the successful cases in the next paragraph. Fig. \ref{ex}.C (a) and Fig. \ref{ex}.C (b) show the position of surfacing, in the simulations and in the real experiment. Comparing them shows that the real environment has a larger distribution of surfacing positions than the simulated environment, even after adding random noise to the simulated dynamics. On the other hand, comparing (b) and (c), indicates that the variance of the surfacing position and distance from the starting position are smaller for the proposed method than for the baseline controller. 

Fig. \ref{ex}.D shows the action values of the functioning actuator in the successful trials (RR, RL, FL) and unsuccessful (FR, RL, FL) trials for the first 15 seconds. Visualized data are derived from the first set of experiments. Fig. \ref{ex}.D indicates that the zero-direction of the oscillations is oriented toward the water surface in both scenarios. On the other hand, the amplitudes in the failed experiments were generally lower than those in the successful ones. This observation shows that the action policies are not symmetrically converging for each actuator. Therefore, it can be considered that the specific allocation failed to surface because of insufficient thrust.

\section{Conclusion}
In this study, we proposed a novel RL based fault-tolerant surfacing controller for underwater robots that operate without requiring explicit identification of malfunctioning actuators. This approach enables the robot to manage a wide range of actuator failures and configurations by learning a robust control policy. A key advantage of our method is that it eliminates the need to identify which actuators have failed, reducing the complexity of the control process and allowing the robot to use operational actuators to complete its task.
Furthermore, Cross-Platform transfer learning was used to share parts of the control policy between different robot designs. This transfer accelerated learning for new robots by reusing knowledge from previously trained models. As shown in Table \ref{tab:tr}, first layer transfers contributed to efficiency in all combinations of platforms compared to models without transfer learning. On the other hand, both first and second layer transfers did not always contribute to efficiency and sometimes decreased it. These results can be  due to the need for weights to reflect the individual characteristics of the actuators in the lower layers. By connecting the encoder whose dynamics is learned with features compressed from the same number of dimensions, it can be possible to design a more general controller with a combination of RL \cite{koop}.

Real-world experiments showed that the RL-based controller achieved a success rate of 85.7 percent (24/28 trials) in real-world tests, outperforming a baseline PID controller, which only achieved a 57.1 percent success rate (8/14 trials). The proposed RL controller exhibited greater stability in maintaining the robot’s attitude, which contributed to the higher success rate. 
This research demonstrates the feasibility of applying simulation-trained RL controllers to real-world underwater robots and provides a scalable solution for various actuator failure scenarios. In future work, we plan to enhance the balance between actuator contributions and extend the methodology to address a wider variety of malfunctions. We also plan to develop general policy functions that do not require additional learning on individual platforms.

\appendix
All simulation code using Gazebo \cite{gazebo} and ROS2 \cite{humble}, dynamics, robot models, and pre-train models are available as open-source project here: \url{https://github.com/Centre-for-Biorobotics/eeUVsim_Gazebo}

\end{document}